%% file: main_Paper.tex
\documentclass[runningheads]{config/llncs}

\input{config/packages}
\input{config/commands}

\begin{document}

\pagestyle{headings}
\mainmatter
\def\ECCVSubNumber{222}

\title{\outTitleFULL}



\titlerunning{\outTitleFULL}
%
\author{Akshita Mittel\inst{1}\index{Mittel , Akshita} \and
Shashank Tripathi\inst{2}\index{Tripathi, Shashank}}
\authorrunning{A. Mittel and S. Tripathi}
%
\institute{NVIDIA \\
\email{amittel@nvidia.com}
\and
Max Planck Institute for Intelligent Systems, T{\"u}bingen, Germany \\
\email{stripathi@tue.mpg.de}}

\maketitle
\subfile{sections_Paper/0_abstract}

\section{Introduction}
\label{sec:intro}
\subfile{sections_Paper/1_intro}

\section{Related Work}
\label{sec:related_work}
\subfile{sections_Paper/2_related}

\section{Method}
\label{sec:method}
\subfile{sections_Paper/3_method}

\section{Experiments}
\label{sec:experiments}
\subfile{sections_Paper/4_experiments}

\section{Conclusion}
\label{sec:conclusion}
\subfile{sections_Paper/5_conclusion}

\clearpage
\bibliographystyle{config/splncs04}
\bibliography{config/BIB}
\end{document}

%% file: config/packages.tex
\usepackage{graphicx}

\usepackage{amsmath}
\usepackage{amssymb}
\usepackage{epsfig}
\usepackage{graphicx}
\usepackage[table]{xcolor}
\usepackage{datetime2}
\usepackage{bm}
\usepackage{color,soul}          
\usepackage{microtype}      
\usepackage{dcolumn}        
\usepackage{url}            
\usepackage{booktabs}       
\usepackage{amsfonts}       
\usepackage{nicefrac}       
\usepackage{enumitem}
\usepackage{multirow}
\usepackage{float}
\usepackage{algorithm}
\usepackage[noend]{algpseudocode}
\usepackage{subfiles}
\usepackage{adjustbox}
\usepackage{subcaption}
\usepackage{makecell}
\usepackage[normalem]{ulem}
\usepackage{caption}
\usepackage{nopageno}
\usepackage[flushmargin,para]{footmisc}
\usepackage{pifont}
\usepackage{pifont}
\usepackage{adjustbox}
\usepackage[pagebackref,breaklinks,colorlinks]{hyperref} 

\usepackage[accsupp]{axessibility}  

\usepackage{wrapfig}

\usepackage[capitalize]{cleveref}
\crefname{section}{Sec.}{Secs.}
\Crefname{section}{Section}{Sections}
\Crefname{table}{Table}{Tables}
\crefname{table}{Tab.}{Tabs.}

\usepackage[numbers,sort]{natbib}

\usepackage{physics}

%% file: config/commands.tex
\usepackage{xspace}
\xspaceaddexceptions{]\}}

\newcommand{\nameCOLOR}[1]{\textcolor{black}{#1}} 
\newcommand{\nameMethod}{\nameCOLOR{\mbox{PERI}}\xspace}
\newcommand{\pas}{\mbox{PAS}\xspace}
\newcommand{\contin}{\mbox{Cont-In}\xspace}

\newcommand{\outTitleFULL}{\nameMethod: Part Aware Emotion Recognition \\ In The Wild}

\newcommand{\etal}{\mbox{et al.}\xspace}
\newcommand{\eg}{\mbox{e.g.}\xspace}

\newcommand{\reffig}[1]{\cref{#1}}

\newcommand{\refsec}[1]{\cref{#1}}

\definecolor{lightgray}{gray}{0.97}
\definecolor{lightblue}{rgb}{0.93,0.95,1.0}

\definecolor{GreenColor}{rgb}{0.137,0.573,0.565}
\definecolor{OrangeColor}{rgb}{0.914,0.541,0.0.141}
\definecolor{PurpleColor}{rgb}{0.5,0,0.7}

%% file: sections_Paper/0_abstract.tex
\begin{abstract}
Emotion recognition aims to interpret the emotional states of a person based on various inputs including audio, visual, and textual cues. This paper focuses on emotion recognition using visual features. To leverage the correlation between facial expression and the emotional state of a person, pioneering methods rely primarily on facial features. However, facial features are often unreliable in natural unconstrained scenarios, such as in crowded scenes, as the face lacks pixel resolution and contains artifacts due to occlusion and blur. To address this, methods focusing on \textit{in the wild} emotion recognition exploit full-body person crops as well as the surrounding scene context. While effective, in a bid to use body pose for emotion recognition, such methods fail to realize the potential that facial expressions, when available, offer. Thus, the aim of this paper is two-fold. First, we demonstrate a method, \nameMethod, to leverage both body pose and facial landmarks. We create \textit{part aware spatial} (\pas) images by extracting key regions from the input image using a mask generated from both body pose and facial landmarks. This allows us to exploit body pose in addition to facial context whenever available. Second, to reason from the \pas images, we introduce context infusion (\contin) blocks. These blocks attend to part-specific information, and pass them onto the intermediate features of an emotion recognition network. Our approach is conceptually simple and can be applied to any existing emotion recognition method. We provide our results on the publicly available in the wild EMOTIC dataset. Compared to existing methods, \nameMethod achieves superior performance and leads to significant improvements in the mAP of emotion categories, while decreasing Valence, Arousal and Dominance errors. Importantly, we observe that our method improves performance in both images with fully visible faces as well as in images with occluded or blurred faces.

\end{abstract}

%% file: sections_Paper/1_intro.tex
The objective of emotion recognition is to recognise how people feel. Humans function on a daily basis by interpreting social cues from around them. Lecturers can sense confusion in the class, comedians can sense engagement in their audience, and psychiatrists can sense complex emotional states in their patients. As machines become an integral part of our lives, it is imperative that they understand social cues in order to assist us better. By making machines more aware of context, body language, and facial expressions, we enable them to play a key in role in numerous situations. This includes monitoring critical patients in hospitals, helping psychologists monitor patients they are consulting, detecting engagement in students, analysing fatigue in truck drivers, to name a few. Thus, emotion recognition and social AI have the potential to drive key technological advancements in the future.  

Facial expressions are one of the biggest indicators of how a person feels. Therefore, early work in recognizing emotions focused on detecting and analyzing faces~\cite{7780969, 7583746, ohman1978facial, russell1985multidimensional}. Although rapid strides have been made in this direction, such methods assume availability of well aligned, fully visible and high-resolution face crops~\cite{tumen2017facial, ko2018brief, mehendale2020facial, hu2019video, duncan2016facial, mellouk2020facial, pranav2020facial}. Unfortunately, this assumption does not hold in realistic and unconstrained scenarios such as internet images, crowded scenes, and autonomous driving. In the wild emotion recognition, thus, presents a significant challenge for these methods as face crops tend to be low-resolution, blurred or partially visible due to factors such as subject's distance from the camera, person and camera motion, crowding, person-object occlusion, frame occlusion etc. In this paper, we address in-the-wild emotion recognition by leveraging face, body and scene context in a robust and efficient framework called \underline{P}art-aware \underline{E}motion \underline{R}ecognition \underline{I}n the wild, or \nameMethod.

Research in psychology and affective computing has shown that body pose offers significant cues on how a person feels~\cite{coulson2004attributing, kleinsmith2007recognizing, de2011bodily}. For example, when people are interested in something, they tilt their head forward. When someone is confident, they tend to square their shoulders. Recent methods recognize the importance of body pose for emotion recognition and tackle issues such as facial occlusion and blurring by processing image crops of the entire body~\cite{kosti2020context, Kosti2017EmotionRI, 9607417, zacharatos2014automatic, 8803460, 8945309}. 
Kosti~\etal~\cite{kosti2020context, Kosti2017EmotionRI} expand upon previous work by adding scene context in the mix,  noting that the surrounding scene plays a key role in deciphering the emotional state of an individual. An illustrative example of this could be of a person crying at a celebration such as graduation as opposed to a person at a funeral. Both individuals can have identical posture but may feel vastly different set of emotions. Huang~\etal ~\cite{9607417} expanded on this by improving emotion recognition using body pose estimations. 

In a bid to exploit full body crops, body keypoints and scene context, such methods tend to ignore part-specific information such as shoulder position, head tilt, facial expressions, etc. which, when available, serve as powerful indicators of the emotional state. While previous approaches focus on either body pose or facial expression, we hypothesize that a flexible architecture capable of leveraging both body and facial features is needed. Such an architecture should be robust to lack of reliable features on both occluded/blurred body or face, attend to relevant body parts and be extensible enough to include context from the scene. To this end, we present a novel representation, called part-aware spatial (\pas) images that encodes both facial and part specific features and retains pixel-alignment relative to the input image. Given a person crop, we generate a part-aware mask by fitting Gaussian functions to the detected face and body landmarks. Each Gaussian in the part-aware mask represents the spatial context around body and face regions and specifies key regions in the image the network should attend to. We apply the part-aware mask on the input image which gives us the final \pas image (see \reffig{figure:context_block}). The \pas images are agnostic to occlusion and blur and take into account both body and face features.   

To reason from \pas images, we propose novel context-infusion (\contin) blocks to inject part-aware features at multiple depths in a deep feature backbone network. Since the \pas images are pixel-aligned, each \contin block implements explicit attention on part-specific features from the input image.  We show that as opposed to \textit{early fusion} (\eg channel-wise concatenation) of \pas image with input image \textbf{I}, or \textit{late fusion} (concatenating the features extracted from \pas images just before the final classification),  \contin blocks effectively utilize part-aware features from the image. \contin blocks do not alter the architecture of the base network, thereby allowing Imagenet pretraining on all layers. The \contin blocks are designed to be easy to implement, efficient to compute and can be easily integrated with any emotion recognition network with minimal effort. 

Closest to our work is the approach of Gunes~\etal~\cite{Gunes2007BimodalER} which combines visual channels from face and upper body gestures for emotion recognition. However, unlike \nameMethod, which takes unconstrained in the wild monocular images as input, their approach takes two high-resolution camera streams, one focusing only on the face and other focusing only on the upper body gestures from the waist up. All of the training data in \cite{Gunes2007BimodalER} is recorded in an indoor setting with a uniform background, single subject, consistent lighting, front-facing camera and fully visible face and body; a setting considerably simpler than our goal of emotion recognition in real-world scenarios. Further, our architecture and training scheme is fundamentally different and efficiently captures part-aware features from monocular images. 

In summary, we make the following contributions:
\begin{enumerate}
    \item Our approach, \nameMethod, advances in the wild emotion recognition by introducing a novel representation (called \pas images) which efficiently combines body pose and facial landmarks such that they can supplement one another. 
    \item We propose context infusion (\contin) blocks which modulate intermediate features of a base emotion recognition network, helping in reasoning from both body poses and facial landmarks. Notably, \contin blocks are compatible with any exiting emotion recognition network with minimal effort. 
    \item Our approach results in significant improvements compared to existing approaches in the publicly-available in the wild EMOTIC dataset~\cite{Kosti2017EMOTICEI}. We show that \nameMethod adds robustness under occlusion, blur and low-resolution input crops.
\end{enumerate}

%% file: sections_Paper/2_related.tex
Emotion recognition is a field of research with the objective of interpreting a person's emotions using various cues such as audio, visual, and textual inputs. Preliminary methods focused on recognising six basic discrete emotions defined by the psychologists Ekman and Friesen~\cite{ekman1971constants}. These include anger, surprise, disgust, enjoyment, fear, and sadness. As research progressed, datasets, such as the EMOTIC dataset~\cite{Kosti2017EMOTICEI,kosti2020context, Kosti2017EmotionRI}, have expanded on these to provide a wider label set. A second class of emotion recognition methods focus not on the discrete classes but on a continuous set of labels described by Mehrabian~\cite{Mehrabian1995FrameworkFA} including Valence (V), Arousal (A), and Dominance (D).
We evaluate the performance of our model using both the 26 discrete classes in the EMOTIC dataset~\cite{Kosti2017EMOTICEI}, as well as valence, arousal, and dominance errors. Our method works on visual cues, more specifically on images and body crops.

\textbf{Emotion recognition using facial features}. Most existing methods in Computer Vision for emotion recognition focus on facial expression analysis ~\cite{7780969, 7583746, ohman1978facial, russell1985multidimensional}. Initial work in this field was based on using the Facial Action Coding System (FACS) ~\cite{Ekman1978FacialAC,  7442563, article2, 5346254, 6967830} to recognise the basic set of emotions. FACS refers to a set of facial muscle movements that correspond to a displayed emotion, for instance raising the inner eyebrow can be considered as a unit of FACS. These methods first extract facial landmarks from a face, which are then used to create facial action units, a combination of which are used to recognise the emotion.  
Another class of methods use CNNs to recognize the emotions~\cite{7780969, Kosti2017EMOTICEI, kosti2020context, 9607417, Mittal2020EmotiConCM, Zhang2019ContextAwareAG}. For instance, Emotionnet~\cite{7780969} uses face detector to obtain face crops which are then passed into a CNN to get the emotion category.
Similar to these methods, we use facial landmarks in our work. However, uniquely, the landmarks are used to create the \pas contextual images, which in turn modulate the main network through a series of convolutions layers in the Cont-In blocks.

\textbf{Emotion recognition using body poses}. Unlike facial emotion recognition, the work on emotion recognition using body poses is relatively new. Research in psychology~\cite{Castillo2019WhatDW, deGelder2006TowardsTN, Gross2012EffortShapeAK} suggests that cues from body pose, including features such as hip, shoulder, elbow, pelvis, neck, and trunk can provide significant insight into the emotional state of a person. 
Based on this hypothesis, Crenn~\etal~\cite{7823448} sought to classify body expressions by obtaining low-level features from 3D skeleton sequences. They separate the features into three categories: geometric features, motion features, and fourier features. Based on these low-level features, they calculate meta features (mean and variance), which are sent to the classifier to obtain the final expression labels. Huang~\etal~\cite{https://doi.org/10.48550/arxiv.2010.06362} use a  body pose extractor built on Actional-Structural GCN blocks as an input stream to their model. The other streams in their model extract information from images and body crops based on the architecture of Kosti~\etal~\cite{kosti2020context, Kosti2017EmotionRI}. The output of all the streams are concatenated using a fusion layer before the final classification.
Gunes~\etal~\cite{Gunes2007BimodalER} also uses body gestures. Similar to \nameMethod, they use facial features by combining visual channels from face and upper body gestures. However, their approach takes two high-resolution camera streams, one focusing only on the face and other focusing only on the upper body gestures, making them unsuitable for unconstrained settings.
For our method, we use two forms of body posture information, body crops and body pose detections. Body crops taken from the original input image are passed into one stream of our architecture. The intermediate features of the body stream are then modulated at regular intervals using Cont-In blocks, which derive information from the \pas image based on body pose and facial landmarks.

\textbf{Adding visual context from the entire image}. The most successful methods for emotion recognition in the wild use context from the entire image as opposed to just the body or facial crop. Kosti~\etal~\cite{kosti2020context, Kosti2017EmotionRI} were among the first to explore emotion recognition in the wild using the entire image as context. They introduced the EMOTIC dataset~\cite{Kosti2017EMOTICEI} on which they demonstrated the efficacy of a two-stream architecture where one of the streams is supplied with the entire image while the other is supplied with body crops. Gupta~\etal~\cite{gupta2018attention} also utilise context from the entire image using an image feature extraction stream. Additionally, the facial crops from the original image are passed through three modules; a facial feature extraction stream, an attention block and finally a fusion network. The attention block utilizes the features extracted from the full image to additionally modulate the features of the facial feature extraction stream. However, unlike Kosti~\etal they focus on recognising just the valence of the entire scene. 
Zhang~\etal~\cite{Zhang2019ContextAwareAG} also use context from an image. Their approach uses a Region Proposal Network (RPN) to detect nodes which then form an affective graph. This graph is fed into a Graph Convolutional Network (GCN) similar to Mittal~\etal~\cite{Mittal2020EmotiConCM}. Similar to Kosti ~\etal the second CNN stream in their network extracts the body features.
Lee ~\etal~\cite{9008268} present CAERNet, which consists of two subnetworks. CAERNet is  two-stream network where one stream works with facial crops and the other in which both facial expression and context (background) are extracted. They use an adaptive fusion network in order to fuse the two streams.
Mittal~\etal~\cite{Mittal2020EmotiConCM} take context a step further. Similar to our approach, they use both body crops and facial landmarks. However, akin to Huang~\etal~\cite{https://doi.org/10.48550/arxiv.2010.06362} they pass body crops and facial landmarks as a separate stream. Their architecture consists of three streams. In addition to the body pose and facial landmark stream, the second stream extracts information from the entire image where the body crop of the person has been masked out. The third stream adds modality in one of two ways. They first encode spatio-temporal relationships using a GCN network similar to \cite{Zhang2019ContextAwareAG}, these are then passed through the third stream. The other method uses a CNN which parses depth images in the third stream.
Similar to these methods, \nameMethod maintains two streams, where one of the stream extracts meaningful context from the entire image while the other focuses on the individual person.

%% file: sections_Paper/3_method.tex
\begin{figure}[t]
\centering
\includegraphics[width=\linewidth]{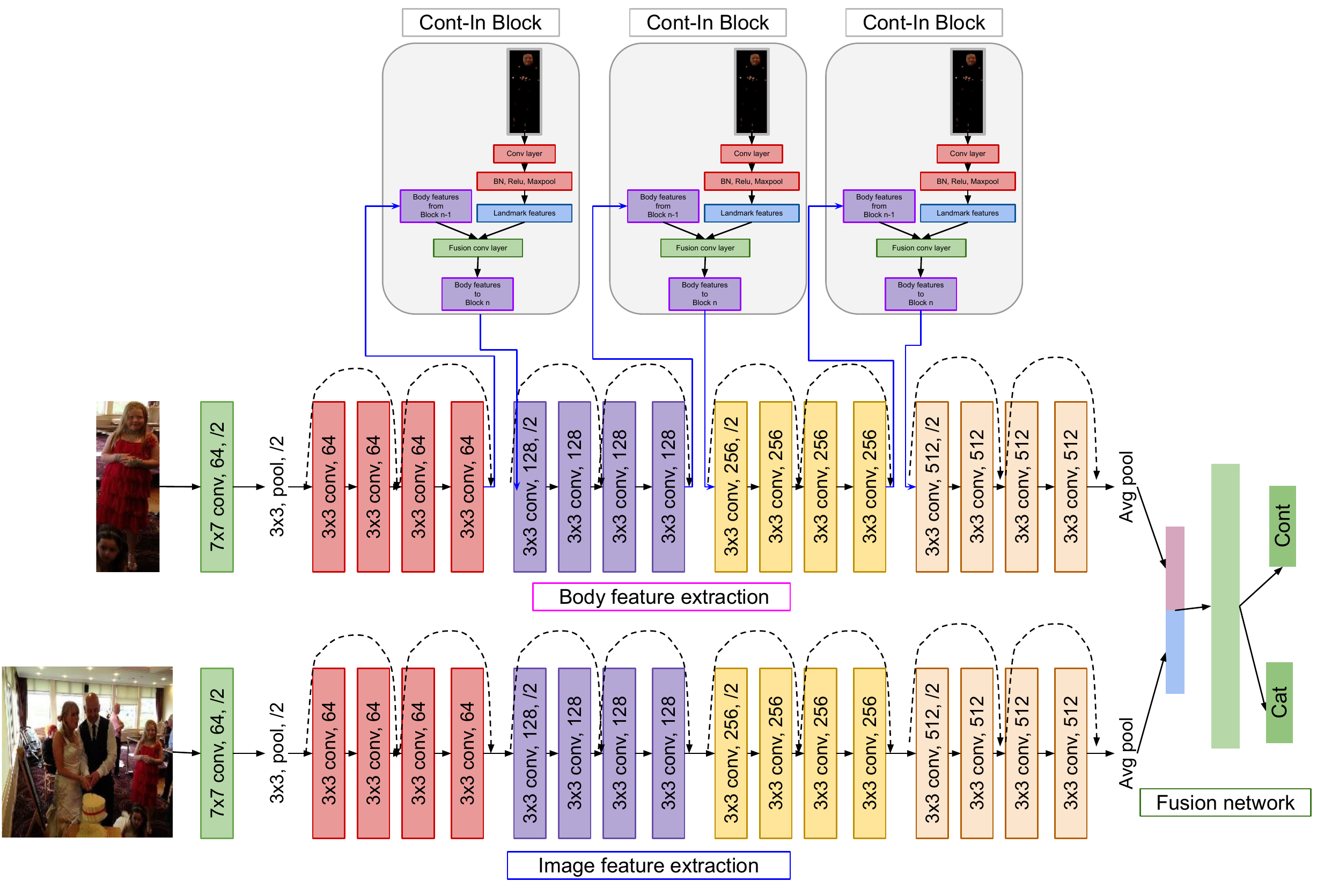}
\caption{
    The overall architecture that consists of a two-stream Resnet-18 network along with the \contin Blocks that modulate features after every intermediate Resnet block using the Part Aware Spatial Context images (PAS). 
}
\label{figure:our_arch}
\end{figure}

The following section describes our framework to effectively recognize emotions from images in the wild. Facial expressions, where available, provide key insight to the emotional state of a person. We find a way to represent body pose and facial landmarks such that we can utilise both set of features subject to their availability in the input image. Concretely, we use MediaPipe’s Holistic model~\cite{DBLP:journals/corr/abs-1906-08172} to obtain landmarks for face and body. These landmarks are then used to create our part aware spatial (\pas) images. Part-specific context from the \pas images is learnt from our context infusion (\contin) blocks which modulate the intermediate features of a emotion detection network. \cref{figure:our_arch} shows the overall framework that we use for our emotion recognition pipeline. A more detailed view of our \contin blocks can be seen in \cref{figure:context_block}.

\subsection{MediaPipe Holistic model}
\label{sec:mediapipe_holistic}
In order to obtain the body poses and facial landmarks, we use the MediaPipe Holistic pipeline~\cite{DBLP:journals/corr/abs-1906-08172}. It is a multi-stage pipeline which includes separate models for body pose and facial landmark detection. The body pose estimation model is trained on $224\times224$ input resolution. However, detecting face and fine-grained facial landmarks requires high resolution inputs. Therefore, the MediaPipe Holistic pipeline first estimates the human pose and then finds the region of interest for the face keypoints detected in the pose output. The region of interest is upsampled and the facial crop is extracted from the original resolution input image and is sent to a separate model for fine-grained facial landmark detection.

\subsection{The Emotic Model}
The baseline of our paper is the the two-stream CNN architecture from Kosti et. al~\cite{kosti2020context, Kosti2017EmotionRI}. The paper defines the task of \textit{emotion recognition in context}, which considers both body pose and scene context for emotion detection. The architecture takes as input the body crop image, which is sent to the body feature extraction stream, and the entire image, which is sent to the image feature extraction stream. The outputs from the two streams are concatenated and combined through linear classification layers. The model outputs classification labels from 26 discrete emotion categories and 3 continuous emotion dimensions, \textit{Valence}, \textit{Arousal} and \textit{Dominance}~\cite{Mehrabian1995FrameworkFA}. The 2 stream architecture is visualized in our pipeline in \cref{figure:our_arch}.  
In order to demonstrate our idea, we stick with the basic Resnet-18~\cite{DBLP:journals/corr/HeZRS15} backbone for both the streams.

\subsection{Part aware spatial image}
\begin{figure}[t]
\centering
\includegraphics[scale=0.5]{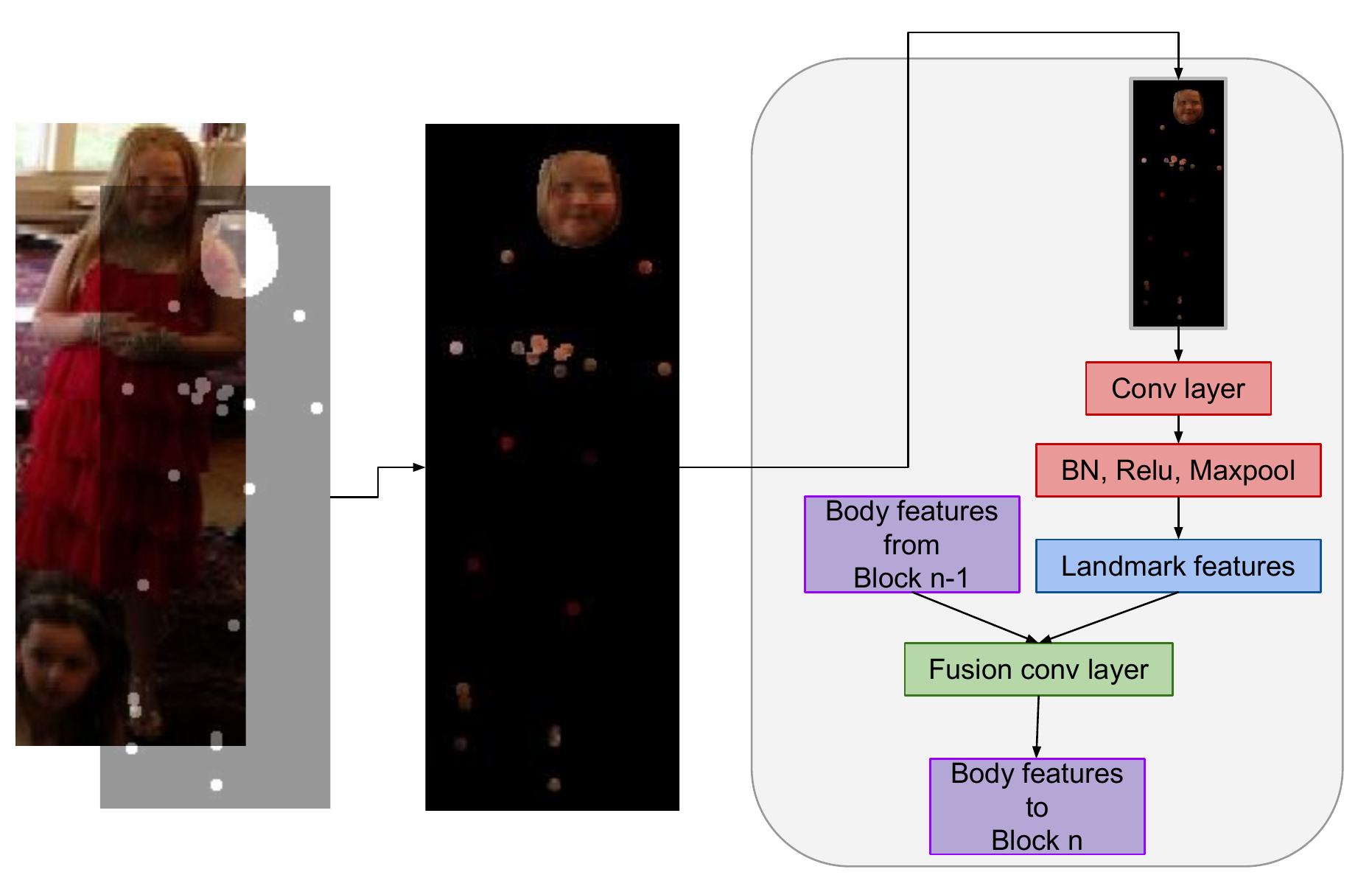}
\caption{
    (Left) An input image along with the mask ($\mathbf{B'}$) created by applying a Gaussian function with $\sigma=3$. The mask is binarised and used to create the PAS image ($\mathbf{P}$) in the middle. (Right) Architecture of the \contin block that uses the PAS images ($\mathbf{P}$) to modulate the Resnet features between each intermediate block. Here the input features from the ${n-1}^{th}$ Resnet block are passed in and modulated features are passed to the $n^{th}$ block are shown in purple.
}

\label{figure:context_block}
\end{figure}
One contribution of our framework is how we combine body pose information along with facial landmarks such that we can leverage both sets of features and allow them to complement each other subject to their availability. In order to do so we have three main stages to our pipeline. First, we use the MediaPipe Holistic model to extract the keypoints as described in \refsec{sec:mediapipe_holistic}. Here we get two sets of keypoint coordinates for each body crop image $\mathbf{I}$. The first set of $N$ coordinates describe the body landmarks $\textbf{b}_i$ where $i \in (0, N)$. The second set of $M$ coordinates describe the location of the facial landmarks $\textbf{f}_j$ where $j \in (0, M)$. For simplicity, we combine all detected landmarks and denote then as $\textbf{b}_k$ where $k \in (0, M+N)$. We take an all black mask $\mathbf{B}\in\mathbb{R}^{1\cp H\cp W}$ the same size as the body crop, and fit a Gaussian kernel to every landmark in the original image as
\begin{equation}
    \mathbf{b}_k'= \frac{1}{\sigma\sqrt{2\pi}}e^{\frac{-(x-\mu)^2}{2\sigma^2}}
    \label{equation:1}
\end{equation}
The part-aware mask $\mathbf{B}'\in\mathbb{R}^{(1\cp H\cp W)}$ is created by binarizing $\mathbf{b}_k'$ using a constant threshold $\rho$, such that
\begin{eqnarray}
    \mathbf{B}'(x) &=& 
    \begin{cases}
        1 & \text{if~}  \| \mathbf{x} - \mathbf{b}_k  \; \|_2 \leq \rho \text{,} \\
        0 & \text{if~}  \| \mathbf{x} - \mathbf{b}_k  \; \|_2 > \rho \text{,}
    \end{cases} 
\end{eqnarray}
where x is the coordinates of all pixels in $\mathbf{B}$. The distance threshold $\rho$ is determined empirically. 

Finally, to obtain the part aware spatial (\pas) image $\mathbf{P}\in\mathbb{R}^{3\cp H\cp W}$, the part-aware mask is applied to the input body crop $\mathbf{I}$ using channel-wise hadamard product, 
\begin{equation}
    \mathbf{P} = \mathbf{I} \otimes \mathbf{B}'
\end{equation}
This process can be visualized in \cref{figure:context_block} (left).

\subsection{Context Infusion Blocks}
To extract information from PAS images, we explore \textit{early fusion}, which simply concatenates PAS with the body crop image $\mathbf{I}$ in the body feature extraction stream of our network. We also explore \textit{late fusion}, concatenating feature maps derived from PAS images before the fusion network. However, both of these approaches failed to improve performance.
Motivated by the above, we present our second contribution, the Context Infusion Block (\contin) which is an architectural block that utilizes the PAS contextual image to condition the base network.  We design \contin blocks such that they can be easily introduced in any existing  emotion recognition network. \cref{figure:context_block} shows the architecture of a \contin block in detail. In \nameMethod, the body feature extraction stream uses the \contin blocks to attend to part-aware context in the input image. Our intuition is that the pixel-aligned \pas images and the \contin block enables the network to determine the body part regions most salient for detecting emotion.  
\contin learns to modulate the network features by fusing the features of the intermediate layer with feature maps derived from PAS. Let $\textbf{X} \in \mathbb{R}^{H\cross W\cross C}$ be the intermediate features from the ${n-1}^{th}$ block of the base network. The PAS image $\textbf{P}$ is first passed through a series of convolutions and activation operations, denoted by $g(.)$, to get an intermediate representation $\mathcal{G} = g(\textbf{P})$ where $\mathcal{G} \in \mathbb{R}^{H\cross W\cross C}$. These feature maps are then concatenated with $\textbf{X}$ to get a fused representation $\textbf{F} = \mathcal{G} \oplus \textbf{X}$. $\textbf{F}$ is then passed through a second series of convolutions, activations, and finally batchnorm to get the feature map $\textbf{X}' \in \mathbb{R}^{H\cross W\cross C'}$ which is then passed through to the $n^{th}$ block of the base network (see \cref{figure:our_arch}).


%% file: sections_Paper/4_experiments.tex
\begin{figure}[t]
\centerline{
\includegraphics[width=\linewidth]{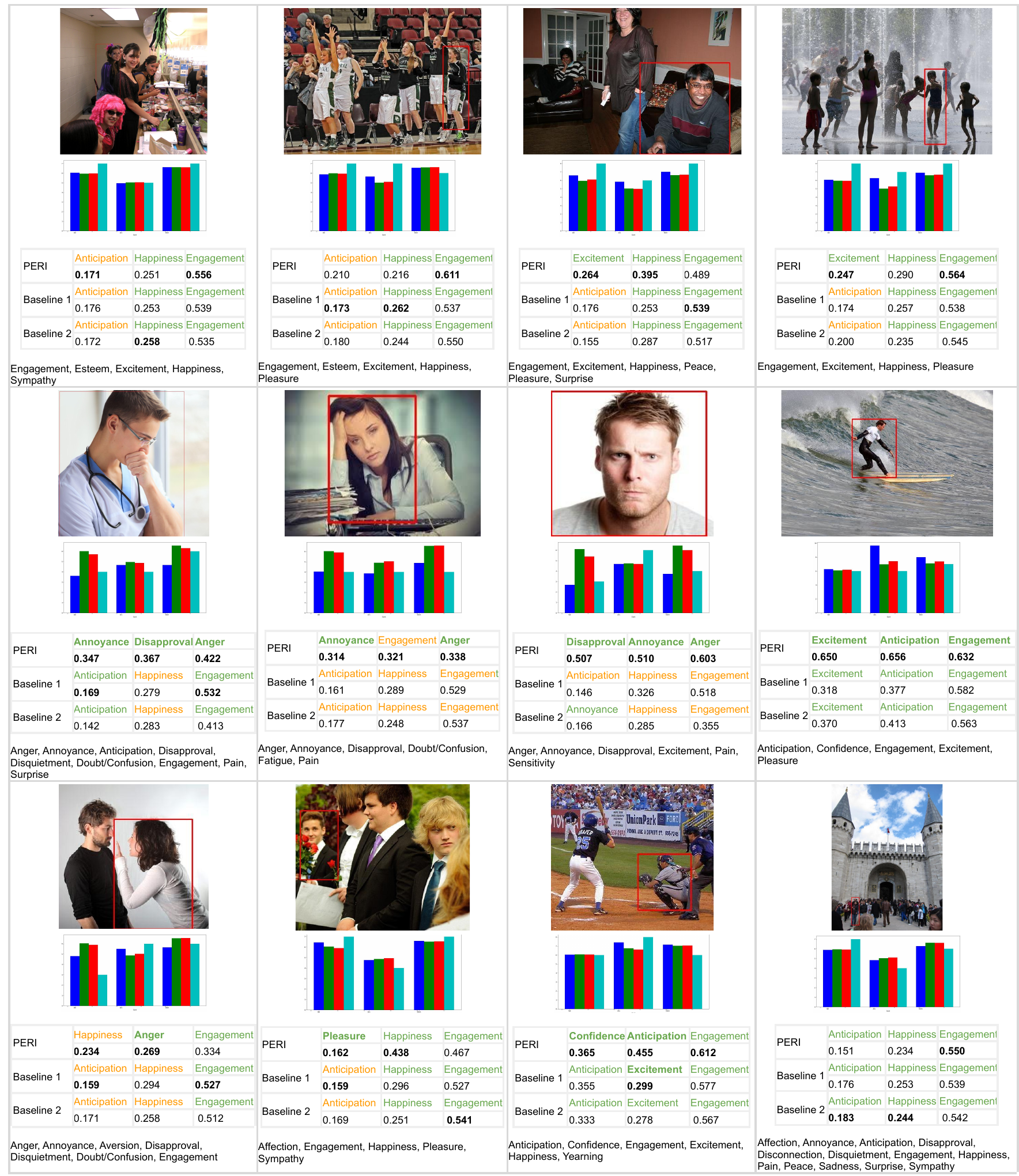}
}
\caption{The figure shows visual results on selected examples. Here each example is separated into 3 parts: the original image with the person of interest in a bounding box; the VAD bar plot (scale 1-10); and the top 3 emotion categories predicted by each model. For VAD values the blue, green, red, cyan columns correspond to \nameMethod, baseline 1 (Kosti~\etal~\cite{kosti2020context, Kosti2017EmotionRI}), baseline 2 (Huang~\etal~\cite{9607417}) and the ground-truth value respectively. For the predicted categories we highlight the category in green if they are present in the ground-truth and orange if they aren't. The ground-truth categories associated with each example are written as a list below the predictions.
}
\label{figure:visual_result}
\end{figure}

\subsection{Experiment setup}
\subsubsection{Dataset and metrics.}
For the purpose of our experiments, we use the two-stream architecture of \cite{kosti2020context, Kosti2017EmotionRI} as our base implementation. 
We use their EMOTIC database~\cite{Kosti2017EMOTICEI}, which is composed of images from MS-COCO~\cite{DBLP:journals/corr/LinMBHPRDZ14}, ADE20K~\cite{zhou2017scene} along with images downloaded from the web.
The database offers two emotion representation labels; a set of 26 discrete emotional categories (Cat), and a set of three emotional dimensions, Valence, Arousal and Dominance from the VAD Emotional State Model~\cite{Mehrabian1995FrameworkFA}.  Valence (V), is a measure of how positive or pleasant an emotion is (negative to positive); Arousal (A), is a measure of the agitation level of the person (non-active, calm, agitated, to ready to act); and Dominance (D) is a measure of the control level of the situation by the person (submissive, non-control, dominant, to in-control). The continuous dimensions (Cont) annotations of VAD are in a 1-10 scale.

\subsubsection{Loss function.}
A dynamic weighted MSE loss $L_{cat}$ is used on the Category classification output layer (Cat) of the model. 
\begin{equation}
L_{cat} = \sum_{i=1}^{26}w_i(\hat{y}_i^{cat} - y_i^{cat})^2
\end{equation}

where $i$ corresponds to 26 discrete categories shown in \cref{table:category}. $\hat{y}_i^{cat}$ and $y_i^{cat}$ are the prediction and ground-truth for the $i^{th}$ category. The dynamic weight $w_i$ are computed per batch and is based on the number of occurrences of each class in the batch. Since the occurrence of a particular class can be 0, \cite{Kosti2017EmotionRI, kosti2020context} defined an additional hyper-parameter $c$. The constant $c$ is added to the dynamic weight $w_i$ along with $p_i$, which is the probability of the $i^{th}$ category. The final weight is defined as $w_i=\frac{1}{ln(p_i+c)}$.

For the continuous (Cont) output layer, an  L1 loss $L_{cont}$ is employed.
\begin{equation}
    L_{cont} = \frac{1}{C}\sum_{i=1}^{C}|\hat{y}_i^{cont} - y_i^{cont}|
\end{equation}
Here $i$ represents one of valence, arousal, and dominance ($C$).
$\hat{y}_i^{cont}$ and $y_i^{cont}$ are the prediction and ground-truth for the $i^{th}$ metric (VAD).

\subsubsection{Baselines.}
We compare \nameMethod to the SOTA baselines including \textit{Emotic} Kosti \etal~\cite{Kosti2017EmotionRI, kosti2020context}, Huang~\etal~\cite{9607417}, Zhang~\etal~\cite{Zhang2019ContextAwareAG}, Lei~\etal~\cite{9008268}, and Mittal~\etal~\cite{Mittal2020EmotiConCM}. We reproduce the three stream architecture in \cite{9607417} based on their proposed method. For a fair comparison, we compare \nameMethod's image-based model results with EmotiCon's~\cite{Mittal2020EmotiConCM} image-based GCN implementation. 

\subsubsection{Implementation details.} We use the two-stream architecture from Kosti \etal~\cite{Kosti2017EmotionRI, kosti2020context}. Here both the image and body feature extraction streams are Resnet-18~\cite{DBLP:journals/corr/HeZRS15} networks which are pre-trained on ImageNet~\cite{ILSVRC15}. All \pas images are re-sized to $128X128$ similar to the input of the body feature extraction stream. The \pas image is created by plotting $501$ landmarks $N+M$ on the base mask and passing it through a Gaussian filter of size $\sigma=3$. We consider the same train, validation, and test splits provided by the EMOTIC~\cite{Kosti2017EMOTICEI} open repository.

\input{TEX_tables/cat}
\input{TEX_tables/VAD}

\subsection{Quantitative results}

\cref{table:category} and \cref{table:vad} show quantitative comparisons between \nameMethod and state-of-the-art approaches. \cref{table:category} compares the average precision (AP) for each discrete emotion category in the EMOTIC dataset~\cite{Kosti2017EMOTICEI}. \cref{table:vad} compares the valence, dominance and arousal $L1$ errors. Our model consistently outperforms existing approaches in both metrics. We achieve a significant $6.3\%$ increase in mean AP (mAP) over our base network~\cite{Kosti2017EmotionRI,kosti2020context} 
and a $1.8\%$ improvement in mAP over the closest competing method~\cite{Mittal2020EmotiConCM}.
Compared to methods that report VAD errors, \nameMethod achieves lower mean and individual $L1$ errors and a $2.6\%$ improvement in VAD error over our baseline~\cite{Kosti2017EmotionRI, kosti2020context}. 
Thus, our results effectively shows that while only using pose or facial landmarks might lead to noisy gradients, especially in images with unreliable/occluded body or face, adding cues from both facial and body pose features where available lead to better emotional context.
We further note that our proposed Cont-In Blocks are effective in reasoning about emotion context when comparing \nameMethod with recent methods that use both body pose and facial landmarks~\cite{Mittal2020EmotiConCM}.

\subsection{Qualitative results}
In order to understand the results further, we look at several visual examples, a subset of which are shown in \cref{figure:visual_result}. 
We choose Kosti~\etal~\cite{Kosti2017EmotionRI,kosti2020context} and Huang~\etal~\cite{9607417} as our baselines as they are the closest SOTA methods. 

We derive several key insights pertaining to our results. In comparison to Kosti~\etal~\cite{Kosti2017EmotionRI,kosti2020context}~\cite{9607417} and Huang~\etal, \nameMethod fares better on examples where the face is clearly visible. This is expected as \nameMethod specifically brings greater attention to facial features. Interestingly, our model also performs better for images where either the face or the body is partially visible (occluded/blurred). This supports our hypothesis that partial body poses as well as partial facial landmarks can supplement one another using our \pas image representation.

\subsection{Ablation study}
As shown in \cref{table:abalation}, we conduct a series of ablation experiments to create an optimal part-aware representation (\pas) and use the information in our base model effectively. For all experiments, we treat the implementation from Kosti~\etal~\cite{Kosti2017EmotionRI, kosti2020context} as our base network and build upon it.

\input{TEX_tables/abalation}

\textbf{\pas images.} To get the best \pas representation, we vary the standard deviation $(\sigma)$ of the Gaussian kernel applied to our \pas image. We show that $\sigma = 3$, gives us the best overall performance with a $5.9\%$ increase in the mAP and a $2.5\%$ decrease in the mean VAD error (\cref{table:abalation}: \pas image experiments) over the base network. From the use of \pas images, we see that retrieving context from input images that are aware of the facial landmarks and body poses is critical to achieving better emotion recognition performance from the base network.


\textbf{Experimenting with Cont-In blocks}. To show the effectiveness of Cont-In blocks, we compare its performance with early and late fusion in \cref{table:abalation}. For early fusion, we concatenate the \pas image as an additional channel to the body-crop image in the body feature extraction stream. For late fusion, we concatenate the fused output of the body and image feature extraction streams with the downsampled \pas image. As opposed to \nameMethod, we see a decline in performance for both mAP and VAD error when considering early and late fusion. From this we conclude that context infusion at intermediate blocks is important for accurate emotion recognition.

Additionally, we considered concatenating the \pas images directly to the intermediate features instead of using a Cont-In block. However, feature concatenation in the intermediate layers changes the backbone ResNet architecture, severely limiting gains from ImageNet~\cite{ILSVRC15} pretraining. This is apparent in the decrease in performance from early fusion, which may be explained, in part, by the inability to load ImageNet weights in the input layer of the backbone network. In contrast, Cont-In block are fully compatible with any emotion recognition network and do not alter the network backbone.    

In the final experiment, we added Cont-In blocks to both the image feature extraction stream and the body feature extraction stream. Here we discovered that if we regulate the intermediate features of both streams as opposed to just the body stream the performance declines. A possible reason could be that contextual information from a single person does generalise well to the entire image with multiple people.
    
\textbf{\nameMethod}. From our ablation experiments, we found that \nameMethod works best overall. It has the highest mAP among the ablation experiments as well as a lowest mean $L1$ error for VAD. While there are other hyper-parameters that have better $L1$ errors for Valence, Arousal, and Dominance independently, (different Gaussian standard deviations ($\sigma_k$)), these hyper-parameters tend to perform worse overall compared to \nameMethod.

%% file: TEX_tables/cat.tex
\begin{table}
\caption{The average precision (AP) results on state-of-the-art methods and \nameMethod. We see a consistent increase across a majority of the discrete class APs as well as the mAP using PERI.}
\centering
\begin{tabular}{l|rrrrrr}
\hline
\textbf{Category}        & \multicolumn{1}{l}{Kosti~\cite{Kosti2017EmotionRI, kosti2020context}} & \multicolumn{1}{l}{Huang~\cite{9607417}} & \multicolumn{1}{l}{Zhang~\cite{Zhang2019ContextAwareAG}} & \multicolumn{1}{l}{Lee~\cite{9008268}} & \multicolumn{1}{l}{Mittal~\cite{Mittal2020EmotiConCM}} & \multicolumn{1}{l}{\textbf{\nameMethod}} \\ \hline
\rowcolor[HTML]{EFEFEF} 
Affection       & 28.06                                                                                 & 26.45                                                    & 46.89                                                                    & 19.90                                                  & 36.78                                                                 & \textbf{38.87}                           \\
Anger           & 6.22                                                                                  & 6.52                                                     & 10.87                                                                    & 11.50                                                  & 14.92                                                                 & \textbf{16.47}                           \\
\rowcolor[HTML]{EFEFEF} 
Annoyance       & 12.06                                                                                 & 13.31                                                    & 11.23                                                                    & 16.40                                                  & 18.45                                                                 & \textbf{20.61}                           \\
Anticipation    & 93.55                                                                                 & 93.31                                                    & 62.64                                                                    & 53.05                                                  & 68.12                                                                 & \textbf{94.70}                           \\
\rowcolor[HTML]{EFEFEF} 
Aversion        & 11.28                                                                                 & 10.21                                                    & 5.93                                                                     & 16.20                                                  & \textbf{16.48}                                                        & 15.55                                    \\
Confidence      & 74.19                                                                                 & 74.47                                                    & 72.49                                                                    & 32.34                                                  & 59.23                                                                 & \textbf{78.92}                           \\
\rowcolor[HTML]{EFEFEF} 
Disapproval     & 13.32                                                                                 & 12.84                                                    & 11.28                                                                    & 16.04                                                  & 21.21                                                                 & \textbf{21.48}                           \\
Disconnection   & 30.07                                                                                 & 30.07                                                    & 26.91                                                                    & 22.80                                                  & 25.17                                                                 & \textbf{36.64}                           \\
\rowcolor[HTML]{EFEFEF} 
Disquietment    & 16.41                                                                                 & 15.12                                                    & 16.94                                                                    & 17.19                                                  & 16.41                                                                 & \textbf{18.46}                           \\
Doubt/Confusion & 15.62                                                                                 & 14.44                                                    & 18.68                                                                    & 28.98                                                  & \textbf{33.15}                                                        & 20.36                                    \\
\rowcolor[HTML]{EFEFEF} 
Embarrassment   & 5.66                                                                                  & 5.24                                                     & 1.94                                                                     & \textbf{15.68}                                         & 11.25                                                                 & 6.00                                     \\
Engagement      & 96.68                                                                                 & 96.41                                                    & 88.56                                                                    & 46.58                                                  & 90.45                                                                 & \textbf{97.93}                           \\
\rowcolor[HTML]{EFEFEF} 
Esteem          & 20.72                                                                                 & 21.31                                                    & 13.33                                                                    & 19.26                                                  & 22.23                                                                 & \textbf{23.55}                           \\
Excitement      & 72.04                                                                                 & 71.42                                                    & 71.89                                                                    & 35.26                                                  & \textbf{82.21}                                                        & 79.21                                    \\
\rowcolor[HTML]{EFEFEF} 
Fatigue         & 7.51                                                                                  & 8.74                                                     & 13.26                                                                    & 13.04                                                  & \textbf{19.15}                                                        & 13.94                                    \\
Fear            & 5.82                                                                                  & 5.76                                                     & 4.21                                                                     & 10.41                                                  & \textbf{11.32}                                                        & 7.86                                     \\
\rowcolor[HTML]{EFEFEF} 
Happiness       & 69.51                                                                                 & 70.73                                                    & 73.26                                                                    & 49.36                                                  & 68.21                                                                 & \textbf{80.68}                           \\
Pain            & 7.23                                                                                  & 7.17                                                     & 6.52                                                                     & 10.36                                                  & 12.54                                                                 & \textbf{16.19}                           \\
\rowcolor[HTML]{EFEFEF} 
Peace           & 21.91                                                                                 & 20.88                                                    & 32.85                                                                    & 16.72                                                  & 35.14                                                                 & \textbf{35.81}                           \\
Pleasure        & 39.81                                                                                 & 40.29                                                    & 57.46                                                                    & 19.47                                                  & \textbf{61.34}                                                        & 49.29                                    \\
\rowcolor[HTML]{EFEFEF} 
Sadness         & 7.60                                                                                  & 8.04                                                     & 25.42                                                                    & 11.45                                                  & \textbf{26.15}                                                        & 18.32                                    \\
Sensitivity     & 5.56                                                                                  & 5.21                                                     & 5.99                                                                     & 10.34                                                  & \textbf{9.21}                                                         & 7.68                                     \\
\rowcolor[HTML]{EFEFEF} 
Suffering       & 6.26                                                                                  & 7.83                                                     & 23.39                                                                    & 11.68                                                  & \textbf{22.81}                                                        & 19.85                                    \\
Surprise        & 11.60                                                                                 & 12.56                                                    & 9.02                                                                     & 10.92                                                  & 14.21                                                                 & \textbf{17.65}                           \\
\rowcolor[HTML]{EFEFEF} 
Sympathy        & 26.34                                                                                 & 26.41                                                    & 17.53                                                                    & 17.13                                                  & 24.63                                                                 & \textbf{36.01}                           \\
Yearning        & 10.83                                                                                 & 10.86                                                    & 10.55                                                                    & 9.79                                                   & 12.23                                                                 & \textbf{15.32}                           \\ \hline
\rowcolor[HTML]{EFEFEF} 
mAP$\uparrow$             & 27.53                                                                                 & 27.52                                                    & 28.42                                                                    & 20.84                                                  & 32.03                                                                 & \textbf{33.86}                          
\end{tabular}
\label{table:category}
\end{table}

%% file: TEX_tables/VAD.tex
\begin{table}
\caption{The VAD and mean errors for continuous labels. The models include the state-of-the-art methods and \nameMethod. We see a consistent decrease across each VAD $L1$ error along with the mean $L1$ error using PERI.}
\centering
\begin{tabular}{@{}l|rrr|r@{}}
\toprule
                                    & \multicolumn{1}{l}{Valence$\downarrow$} & \multicolumn{1}{l}{Arousal$\downarrow$} & \multicolumn{1}{l|}{Dominance$\downarrow$} & \multicolumn{1}{l}{VAD Error$\downarrow$} \\ \midrule
\rowcolor[HTML]{EFEFEF} 
Kosti~\etal~\cite{Kosti2017EmotionRI,kosti2020context}                            & 0.71                      & 0.91                      & 0.89                         & 0.84                       \\
Huang~\etal~\cite{9607417}             & 0.72                      & 0.90                      & 0.88                         & 0.83    
\\
\rowcolor[HTML]{EFEFEF}
Zhang~\etal~\cite{Zhang2019ContextAwareAG}             & \textbf{0.70}                      & 1.00                      & 1.00                         & 0.90                        \\
 
\nameMethod & \textbf{0.70}             & \textbf{0.85}             & \textbf{0.87}                & \textbf{0.80}              
\end{tabular}
\label{table:vad}
\end{table}

%% file: TEX_tables/abalation.tex
\begin{table}[]
\caption{Ablation studies. We divide this table into two sections. 1) Experiments to obtain the best \pas image. 2) Experiments to get the optimum method to use these \pas images (Cont-In blocks)}
\begin{tabular}{lrrrrr}
\hline
\rowcolor[HTML]{EFEFEF} 
\multicolumn{1}{l|}{\cellcolor[HTML]{EFEFEF}}& \multicolumn{1}{l|}{\cellcolor[HTML]{EFEFEF}\textbf{mAP$\uparrow$}} & \multicolumn{1}{l}{\cellcolor[HTML]{EFEFEF}\textbf{Valence$\downarrow$}} & \multicolumn{1}{l}{\cellcolor[HTML]{EFEFEF}\textbf{Arousal$\downarrow$}} & \multicolumn{1}{l|}{\cellcolor[HTML]{EFEFEF}\textbf{Dominance$\downarrow$}} & \multicolumn{1}{l}{\cellcolor[HTML]{EFEFEF}\textbf{Avg error$\downarrow$}} \\ \hline
\multicolumn{6}{c}{\textbf{Baselines}}\\ \hline

\multicolumn{1}{l|}{Kosti~\etal~\cite{Kosti2017EmotionRI,kosti2020context}}& \multicolumn{1}{r|}{27.53}& 71.16& 90.95& \multicolumn{1}{r|}{88.63}& 83.58\\
\rowcolor[HTML]{EFEFEF} 
\multicolumn{1}{l|}{Huang~\etal~\cite{9607417}}& \multicolumn{1}{r|}{27.52}& 72.22& 89.92& \multicolumn{1}{r|}{88.31}& 83.48\\ \hline
\multicolumn{6}{c}{\textbf{PAS image experiments}}\\ \hline
\multicolumn{1}{l|}{PAS $\sigma=1$}& \multicolumn{1}{r|}{33.32}& 70.77& \textbf{83.75}& \multicolumn{1}{r|}{89.47}& 81.33\\
\rowcolor[HTML]{EFEFEF} 
\multicolumn{1}{l|}{\cellcolor[HTML]{EFEFEF}PAS $\sigma=3$}& \multicolumn{1}{r|}{\cellcolor[HTML]{EFEFEF}33.80}& 71.73& 85.36& \multicolumn{1}{r|}{\cellcolor[HTML]{EFEFEF}86.36}& 81.15\\
\multicolumn{1}{l|}{PAS $\sigma=5$}& \multicolumn{1}{r|}{33.46}& \textbf{70.36}& 87.56& \multicolumn{1}{r|}{\textbf{85.39}}& 81.10\\
\rowcolor[HTML]{EFEFEF} 
\multicolumn{1}{l|}{\cellcolor[HTML]{EFEFEF}PAS $\sigma=7$}& \multicolumn{1}{r|}{\cellcolor[HTML]{EFEFEF}32.74}& 70.95& 85.46& \multicolumn{1}{r|}{\cellcolor[HTML]{EFEFEF}88.04}& 81.48\\
 \hline
\multicolumn{6}{c}{\textbf{Cont-In block experiments}}\\ \hline
\rowcolor[HTML]{EFEFEF}
\multicolumn{1}{l|}{Early fusion}& \multicolumn{1}{r|}{32.96}& 70.60& 85.95& \multicolumn{1}{r|}{87.59}& 81.38\\

\multicolumn{1}{l|}{Late fusion}& \multicolumn{1}{r|}{32.35}& 71.73& 85.60& \multicolumn{1}{r|}{87.12}& 81.48\\
\rowcolor[HTML]{EFEFEF} 
\multicolumn{1}{l|}{\cellcolor[HTML]{EFEFEF}Cont-In on both streams} & \multicolumn{1}{r|}{\cellcolor[HTML]{EFEFEF}29.30}& 72.43& 87.23& \multicolumn{1}{r|}{\cellcolor[HTML]{EFEFEF}8997}& 83.21\\
\multicolumn{1}{l|}{PERI}& \multicolumn{1}{r|}{\textbf{33.86}}& 70.77& 84.56& \multicolumn{1}{r|}{87.49}& \textbf{80.94}\\
\end{tabular}

\label{table:abalation}
\end{table}

%% file: sections_Paper/5_conclusion.tex
Existing methods for in the wild emotion recognition primarily focus on either face or body, resulting in failures under challenging scenarios such as low resolution, occlusion, blur etc. To address these issues, we introduce \nameMethod, a method that effectively represents body poses and facial landmarks together in a pixel-aligned part aware contextual image representation, \pas. We argue that using both features results in complementary information which is effective in challenging scenarios. Consequently, we show that \pas allows better emotion recognition not just in examples with fully visible face and body features, but also when one of the two features sets are missing, unreliable or partially available. 

To seamlessly integrate the \pas images with a baseline emotion recognition network, we introduce a novel method for modulating intermediate features with the part-aware spatial (\pas) context by using context infusion (Cont-In) blocks.  We  demonstrate that using Cont-In blocks works better than a simple early or late fusion. 
\nameMethod significantly outperforms the baseline emotion recognition network of Kosti~\etal~\cite{Kosti2017EMOTICEI, Kosti2017EmotionRI}. \nameMethod also improves upon existing state-of-the-art methods on both the mAP and VAD error metrics.  

While our method is robust towards multiple in-the-wild challenging scenarios, we do not model multiple-human scenes and dynamic environments. In the future, we wish to further extend Cont-In blocks to utilise the \pas context better. Instead of modeling explicit attention using \pas images, it might be interesting to learn part-attention implicitly using self and cross-attention blocks, but we leave this for future work. Additionally, we also seek to explore multi-modal input beyond images, such as depth, text and audio.